\pdfoutput=1

\documentclass[11pt]{article}

\usepackage{coling}

\usepackage{times}
\usepackage{latexsym}
\usepackage[T1]{fontenc}
\usepackage[utf8]{inputenc}
\usepackage{microtype}
\usepackage{inconsolata}
\usepackage{graphicx}

\usepackage{amsmath}
\usepackage{booktabs}
\usepackage[frozencache,cachedir=.]{minted}
\setminted{fontsize=\small}

\title{Reliable, Reproducible, and Really Fast Leaderboards with Evalica}

\author{Dmitry Ustalov \\
  JetBrains / Belgrade, Serbia \\
  \texttt{dmitry.ustalov@jetbrains.com}}

\begin{document}

\maketitle

\begin{center}
\includegraphics[width=10em]{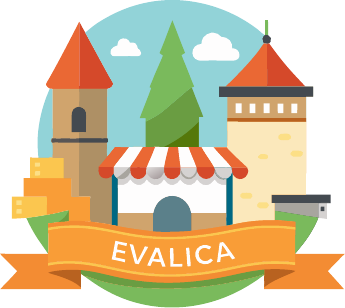}\medskip
\end{center}

\begin{abstract}
The rapid advancement of natural language processing (NLP) technologies, such as instruction-tuned large language models (LLMs), urges the development of modern evaluation protocols with human and machine feedback. We introduce \textbf{Evalica}, an open-source toolkit that facilitates the creation of reliable and reproducible model leaderboards. This paper presents its design, evaluates its performance, and demonstrates its usability through its Web interface, command-line interface, and Python API.\medskip
\end{abstract}

%

\section{\label{sec:intro}Introduction}

\begin{figure}[t]
  \centering
  \includegraphics[scale=.3]{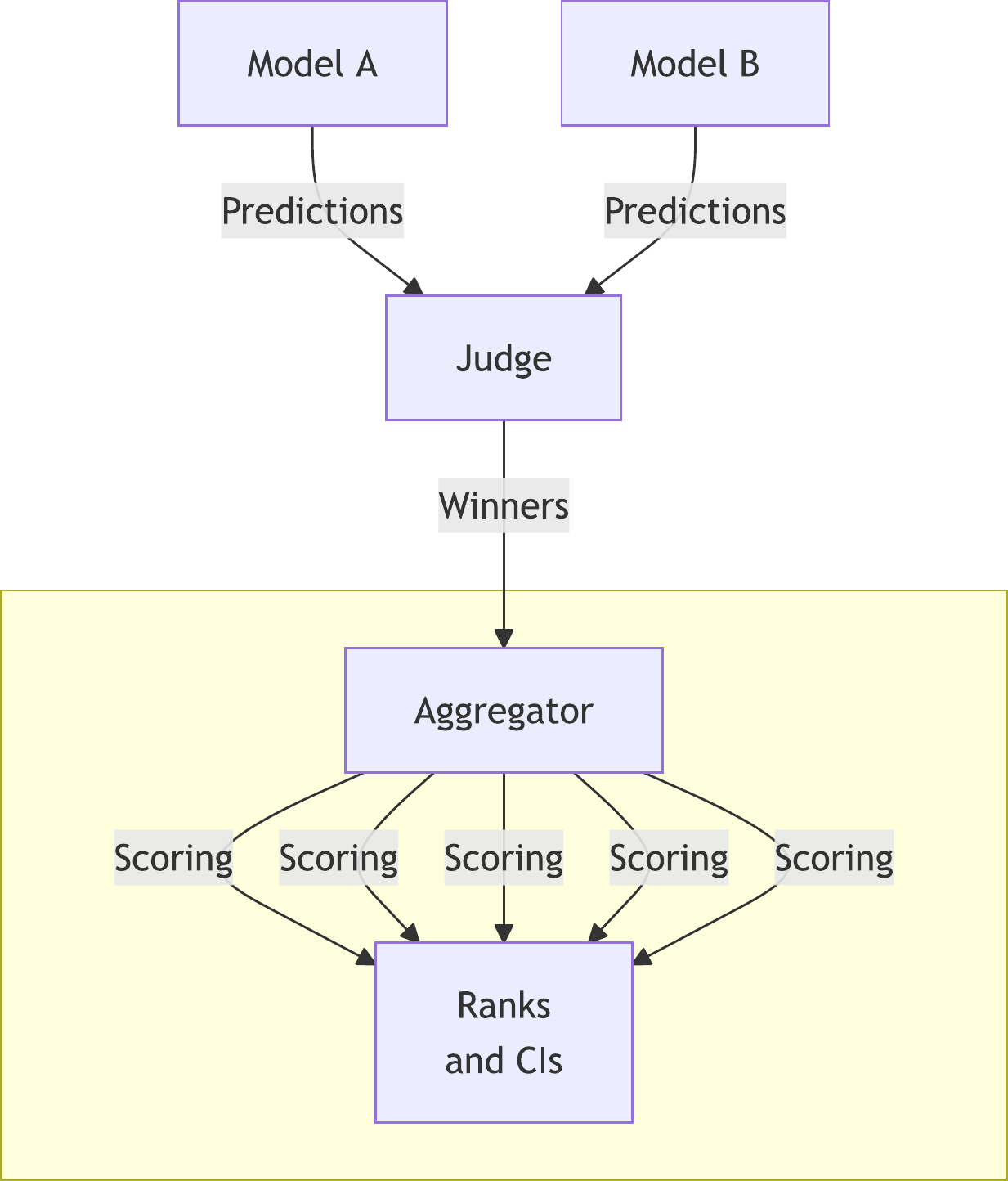}
  \caption{\label{fig:approach}Evalica facilitates the highlighted aspects of leaderboard-making that involve aggregation of judgements, scoring the models with bootstrapped confidence intervals (CIs), and getting the final model ranks.}
\end{figure}

The emergent abilities, as exhibited by highly capable natural language processing (NLP) methods, such as instruction-tuned large language models (LLMs), urge the development of sound and reliable evaluation protocols. While the earlier methods could be reasonably evaluated on static datasets or individual benchmarks, modern methods require up-to-date benchmarks with live feedback from humans and machines \citep{Faggioli:24}. These benchmarks are often represented as pairwise comparison leaderboards (Figure~\ref{fig:approach}), as popularized by LMSYS~Arena \citep{Chiang:24} and AlpacaEval \citep{Dubois:24} projects.

As the NLP methodology evolves rapidly, today's evaluation methods are often implemented in computational notebooks and ad-hoc programs as an afterthought, which introduces errors, incompatibilities, and harms reproducibility and adoption. To improve the engineering aspect of benchmarking by reducing the number of methodological errors and simplifying the exchange and interpretation of the results, we present \textbf{Evalica}, an open-source evaluation toolkit that facilitates and speeds up the creation of reliable and reproducible NLP model benchmarks,\footnote{\url{https://github.com/dustalov/evalica}}  currently focused on the preference data. Based on our four-year experience in the development of production-grade tooling for quality control in crowdsourcing \citep{Ustalov:24:crowdkit}, we built Evalica with three practical goals in mind:
\begin{itemize}
  \item \emph{make the popular evaluation practices available} for a wide audience of users
  \item \emph{ensure the performance and correctness} of the offered implementations
  \item \emph{provide the best developer experience} possible
\end{itemize}

The remainder of this paper is organized as follows. Section~\ref{sec:related} reviews the related work and its relationship with the declared goals. Section~\ref{sec:design} shows Evalica's design and how it satisfies these goals. Section~\ref{sec:implementation} describes the technical details of the Evalica implementation, including the means to ensure its correctness. Section~\ref{sec:performance} reports performance benchmarks against alternative implementations. Finally, Appendix~\ref{sec:usage} demonstrates a Web, a command-line , and a Python application programming interfaces (API) of Evalica.

\section{\label{sec:related}Related Work}

The research community has been developing various toolkits for ranking systems, such as \citet{Elo:78} and TrueSkill \citep{Herbrich:06}. In our analysis, we distinguish several classes of them.

First, \emph{dedicated leaderboard building tools}, such as IFEval~\cite{Zhou:23}, LMSYS~Arena \citep{Chiang:24}, Arena-Hard \citep{Li:24}, and AlpacaEval \citep{Dubois:24}. These toolkits were created by teams of researchers to implement a specific novel evaluation methodology. The code was generally written strictly tailored to the particular benchmark, requiring extra effort from the user to apply it to their own dataset and domain. Due to the high pace of today's scientific research, certain software engineering best practices were often omitted, such as test coverage, code documentation, continuous integration, and data format compatibility. At the same time, some implementations suffer from suboptimal computational performance on larger realistic datasets, which were out of scope of the original benchmarks.

Second, \emph{ranking system implementations}, including Rust packages Propagon\footnote{\url{https://github.com/Refefer/propagon}} and skillrating,\footnote{\url{https://github.com/atomflunder/skillratings}} a Python package OpenSkill.py \citep{Joshi:24}, and others. As these packages are often written by skilled programmers in the best effort to bring correct implementations, these methods do not always match the ones used in current best practices in NLP evaluation. Also, the non-Python packages require an additional non-trivial effort to integrate with the existing Python code and notebooks.

Finally, \emph{application-specific toolkits} like Elovation,\footnote{\url{https://github.com/elovation/elovation}} ArtistAssistApp,\footnote{\url{https://github.com/eugene-khyst/pairwise-comparison}} and Crowd-Kit \citep{Ustalov:24:crowdkit}. These toolkits were built to accommodate user-generated content, usually in the form of crowdsourcing annotation, and often do not follow the methodology used in NLP evaluation.

\section{\label{sec:design}Design of Evalica}

\begin{figure}[t]
  \centering
  \includegraphics[width=\columnwidth]{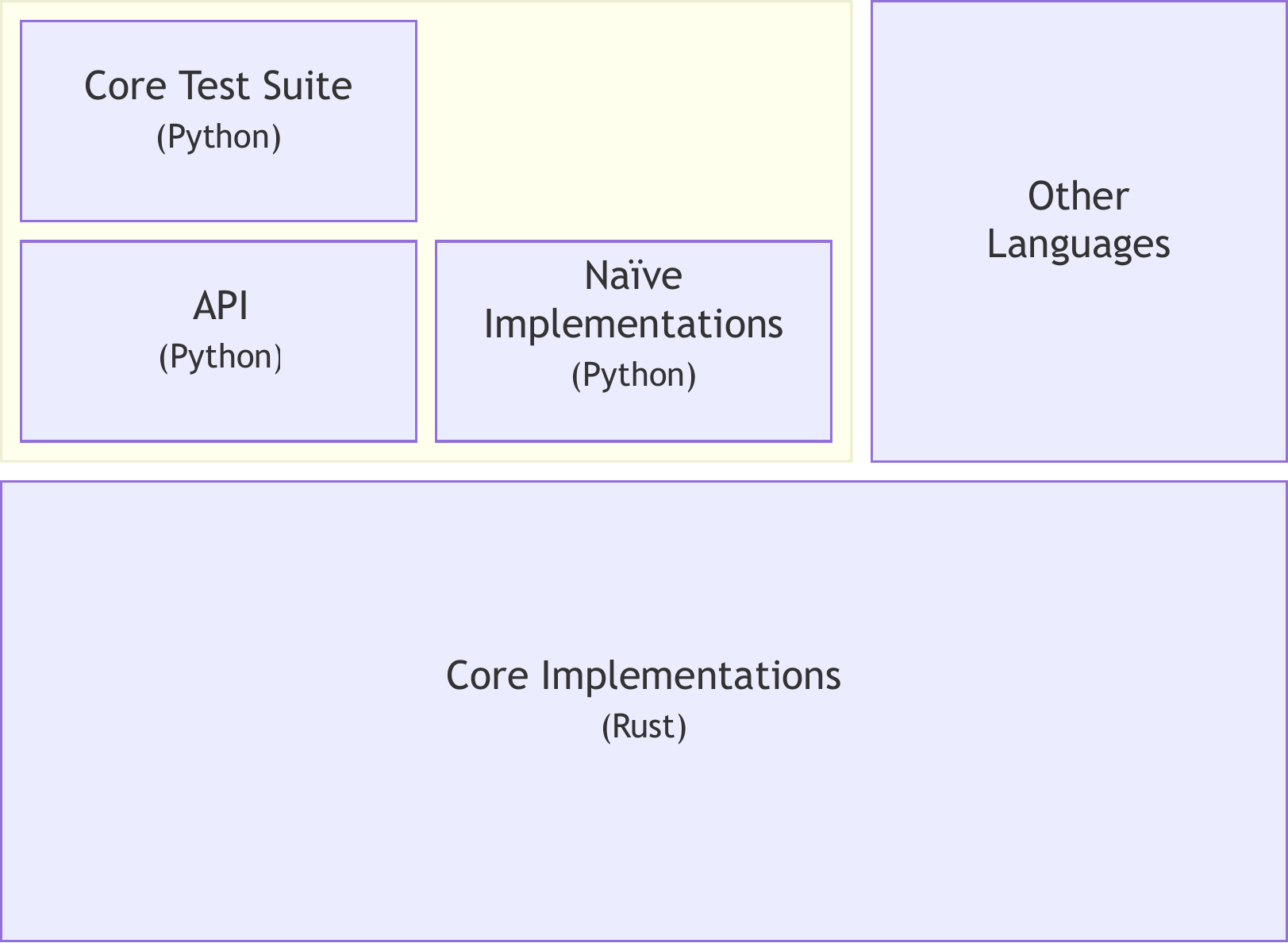}
  \caption{\label{fig:architecture}Evalica has a core in Rust that is covered by a comprehensive suite of tests in Python. We simplify prototyping and increase test reliability by keeping an independent implementation of each method in Python.}
\end{figure}

Evalica facilitates three tasks shown in Figure~\ref{fig:approach}: it provides optimized single-threaded implementations of rating systems, simplifies the computation of confidence intervals for model scores, and offers convenient routines to prepare visualizations.

Figure~\ref{fig:architecture} outlines the architecture of Evalica. In its \emph{core}, there are performance-critical routines in Rust that process the raw data. These core routines are wrapped in convenient APIs for application developers in other languages. These APIs were responsible for transforming the representation into the indexed format as used by the core routines.\footnote{Models usually have names like \texttt{llama-3.1-405b-instruct} and \texttt{claude-3-5-sonnet-20240620}. Computers do not operate with strings \emph{per se}, so we need to transform such names into the corresponding indices, e.g., \texttt{0} and \texttt{1}.} Examples of core routines are all the ranking algorithm implementations and helper routines for constructing win matrices.

We currently only support Python due to its popularity in machine learning. For the sake of reliability and ease of prototyping, we na\"{i}vely and implemented all the methods additionally in Python and built a comprehensive test suite that compares the Python implementations with the Rust ones. Other languages can be supported relatively easily (as long as there exists a bridge between Rust and that language), and improvements to the core implementations and tests will improve the state of all the derivative code.

We believe that these measures allowed satisfying the three goals mentioned in Section~\ref{sec:intro} adequately. Evalica accelerates popular evaluation practices by shipping the corresponding implementations in a high-performing compiled programming language, building on lessons learned from previously developed software to increase the developer's productivity.

\section{\label{sec:implementation}Implementation Details}

Evalica implements scoring approaches from popular benchmarks, such as Chatbot Arena and Arena-Hard: \citet{Elo:78} and \citet{Bradley:52}, and average win rate. We ensured that they provided the same results as in these benchmarks. The package also contains implementations of the eigenvalue method \citep{Bonacich:87}, PageRank \citep{Brin:98}, tie-aware method of \citet{Newman:23}, and trivial vote counting.

To invoke the implementation in Evalica (Listing~\ref{lst:elo}), one needs to supply a vector of left objects (\texttt{xs}), a vector of right objects (\texttt{ys}), a vector of winner labels (\texttt{winners}), and, optionally, a vector of example weights (\texttt{weights}) for style control, as proposed in \citet{Li:24}. Possible values of the winner labels are ``X won,'' ``Y won,'' and ``tie.'' All methods are available in a lightweight and uniform functional API. We intentionally decided to avoid making assumptions about the tabular form of data as our experience in running Crowd-Kit \citep{Ustalov:24:crowdkit} in production showed that it required an error-prone data transformation step that could have been avoided.

Internally, Evalica does not operate with model names, and core implementations require an index to compare the model name to the unique numerical identifier (as described in Section~\ref{sec:design}). Since this operation takes short yet non-negligible time, we provided the possibility to pass the already built index to save time during bootstrapping the confidence intervals and other routines that require resampling and recomputing the scores (Listing~\ref{lst:bootstrap}).

Besides the API, Evalica offers a built-in Web interface and a command-line interface, see Appendix~\ref{sec:usage} for illustrative examples. More specifically, the built-in Web interface follows a well-known input-output separation paradigm from \citet{Abid:19} and was created using the Gradio toolkit (Figure~\ref{fig:gradio}).\footnote{\url{https://www.gradio.app/}} The command-line interface was developed using the pandas library for data manipulation \citep{McKinney:10} and the tools available from the Python standard library (Figure~\ref{fig:cli}).

After computing the scores and ranks, it is often useful to visualize the pairwise win rates for the compared models. Following \citet{Chiang:24}, we applied the \citet{Bradley:52} definition of such a quantity for all pairs of models $i$ and $j$:
\begin{equation*}
  p_{ij} = \frac{s_i}{s_i + s_j}\text{,}
\end{equation*}
where $p_{ij}$ is the probability of model $i$ winning against the model $j$, $s_i$ is the score of model $i$, and $s_j$ is the score of model $j$.

\subsection{Correctness and Reliability}

We applied a set of reasonable means to ensure correctness and reliability of the method implementations in Evalica. First, we implemented all the methods independently in two different programming languages, Rust and Python. We ensured that the outputs for the same inputs are the same between these implementations. Second, we employed property-based tests with the Hypothesis library \citep{MacIver:19} for Python, which enumerated corner cases including empty or illegal inputs to break the program.\footnote{\url{https://github.com/HypothesisWorks/hypothesis}} We covered all such cases and provide reasonable numerical fallbacks, where possible. Third, we compared the outputs against the canonical scores from external benchmarks. Fourth, we ensured that the test coverage is no less than 100\%, and the test suite was executed on every revision in the repository.

\subsection{Governance and Availability}

We built Evalica using the trusted open-source ecosystem. The source code of Evalica was available under the Apache License 2.0 on GitHub.\footnote{\url{https://github.com/dustalov/evalica}} Feature requests and code contributions were processed using the Issues and Pull Requests features on GitHub, correspondingly. We used continuous integration on GitHub Actions to invoke per-revision checks, including unit tests, linting, type checking, test coverage measurement, and computational performance testing. Public dashboards with test coverage and performance tests were available on Codecov\footnote{\url{https://codecov.io/gh/dustalov/evalica}} and Codspeed,\footnote{\url{https://codspeed.io/dustalov/evalica}} correspondingly. We used the trusted publishing approach to release Python packages to PyPI for the Linux, Windows, and macOS platforms.\footnote{\url{https://pypi.python.org/pypi/evalica}} Our compiled packages were forward compatible with any version of Python newer than 3.8 due to the use of the stable CPython ABI. We also released Evalica on conda-forge for the users of Anaconda, a popular distribution of scientific computing tools.\footnote{\url{https://anaconda.org/conda-forge/evalica}} Last but not least, we published the developer documentation on Read the Docs.\footnote{\url{https://evalica.readthedocs.io/}}

\section{\label{sec:performance}Performance Tests}

We performed two series of computational experiments to study the running time of algorithm implementations in Evalica after ensuring their correctness. First, we evaluated the difference in computational performance between the current implementations in a popular benchmark and the ones provided by Evalica. Second, we compared the performance of core and na\"{i}ve implementations of all the methods inside Evalica. All the experiments were run using CPython~3.13.1, NumPy~2.2.0, and Evalica~0.3.2 on macOS~15.2 (Intel\textsuperscript{\textregistered{}} Core\textsuperscript{\texttrademark{}} i5-8500 CPU, 32~GB RAM). All confidence intervals were built using bootstrap with 10K samples and 95\% significance level.

\subsection{Chatbot Arena Experiment}

\begin{table}[t]
\centering
\begin{tabular}{lr}\toprule
\textbf{Setup} & \textbf{Time} $\uparrow{}$ \\\midrule
BT in Evalica & $1.174 \pm 0.009$ \\
Elo in Evalica & $1.256 \pm 0.019$ \\
Elo from Arena-Hard & $3.778 \pm 0.322$ \\
BT from Chatbot Arena & $51.949 \pm 1.797$ \\\bottomrule
\end{tabular}
\caption{\label{tab:arena}Performance of Evalica, Chatbot Arena, and Arena-Hard on the Chatbot Arena dataset. Time is in seconds; a 95\% confidence interval is shown for ten runs. Smaller is better. BT means \citet{Bradley:52}, Elo means \citet{Elo:78}.}
\end{table}

We evaluated the performance of four setups in processing the August 14, 2024 version of the Chatbot Arena dataset \citep{Chiang:24} that contained 1.7M pairwise comparisons of 129 models, ties were not excluded.\footnote{\url{https://storage.googleapis.com/arena_external_data/public/clean_battle_20240814_public.json}} We compared four different setups: an implementation of the \citet{Elo:78} ranking system in pure Python, as used in Chatbot Arena, an implementation of \citet{Bradley:52} in Python with scikit-learn \citep{Pedregosa:11}, as used in Arena-Hard, and Rust implementations of these two methods in Evalica. For that, we ran every setup ten times to simulate the realistic problem of  confidence interval estimation that does often appear in model leaderboards. As the results in Table~\ref{tab:arena} indicate, Evalica's implementations of ranking methods outperformed the current ones as used in the benchmarks by up to 46 times without any involvement of multi-threading processing. Although this was expected since Python is an interpreted language and Rust is a compiled language, we believe that the Evalica's combination of performance and ergonomics would allow running more experiments within the same time budget. At the same time, performing computation in multiple threads, e.g., processing one sampling round per thread, would allow one to better use of the modern multi-core CPUs and reduce the computation time by multiple times.

\subsection{Rust vs. Python in Evalica Experiment}

\begin{table}[t]
\centering
\resizebox{\columnwidth}{!}{\begin{tabular}{lrr}\toprule
\textbf{Algorithm} & \textbf{Rust} & \textbf{Python} \\\midrule
Average Win Rate & $0.005 \pm 0.000$ & $0.006 \pm 0.000$ \\
Bradley--Terry & $0.005 \pm 0.000$ & $0.012 \pm 0.000$ \\
Counting & $0.005 \pm 0.000$ & $0.009 \pm 0.000$ \\
Eigenvalue & $0.005 \pm 0.000$ & $0.006 \pm 0.000$ \\
Elo & $0.005 \pm 0.000$ & $0.484 \pm 0.004$ \\
Newman & $0.006 \pm 0.000$ & $0.010 \pm 0.000$ \\
PageRank & $0.005 \pm 0.000$ & $0.006 \pm 0.000$ \\\bottomrule
\end{tabular}}
\caption{\label{tab:evalica}Running time comparison of core Rust and na\"{i}ve Python implementations of methods in Evalica on the LLMFAO dataset. Time is in seconds; a 95\% confidence interval for ten runs is shown for each implementation. Smaller is better.}
\end{table}

\begin{figure*}[t]
  \centering
  \includegraphics[width=\linewidth]{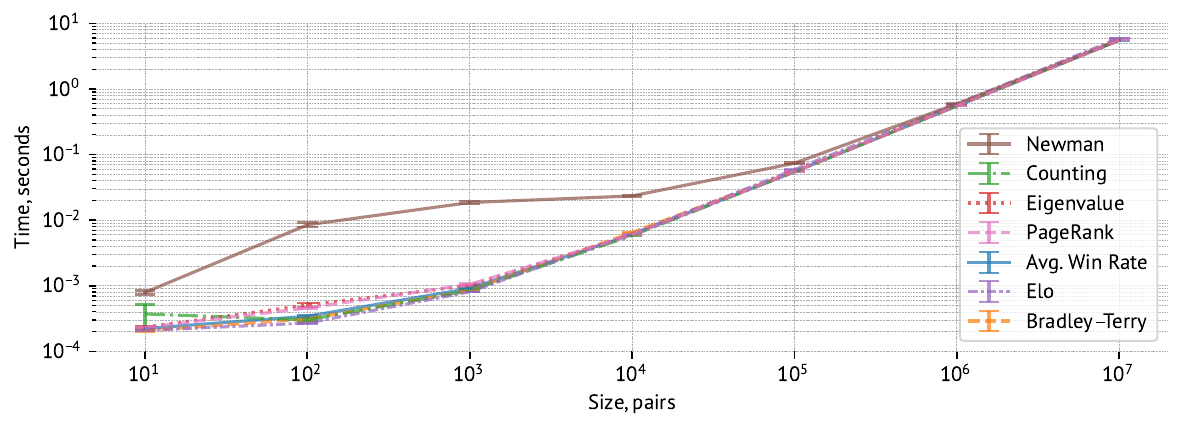}
  \caption{\label{fig:scale}Performance scaling analysis of the Rust implementations in Evalica on the synthetic version of the Chatbot Arena dataset. Both scales are logarithmic. Time is in seconds, dataset size is the number of pairs; a 95\% confidence interval is shown for ten runs. Lower is better.}
\end{figure*}

We evaluated the performance of all the methods implemented in Evalica's core in Rust against their na\"{i}ve implementations in Python. Despite the name, these Python implementations were written using NumPy \citep{Harris:20}, a highly optimized library built on decades of successful performance engineering work for numerical computation in C and Fortran. We used the dataset from a smaller benchmark called LLMFAO \citep{Ustalov:23:llmfao}, which had 9K pairwise comparisons for 59 LLMs, gathered in October 2023 using crowdsourcing. As the results in Table~\ref{tab:evalica} show, the differences between core and na\"{i}ve implementation were statistically significant, according to the permutation test ($p < 0.01$), but the effect size was not noticeable on that scale due to the efficient NumPy routines used in the pure Python implementations. One important exception was Elo, whose equivalent implementation in Rust appeared to be more than 96 times faster than in Python due to the efficient compiler optimizations. At the same time, the Rust implementations had a smaller runtime variance and more predictable performance, which should be useful on larger-scale datasets.

\subsection{Scaling on Synthetic Data Experiment}

We analyzed the relationship between dataset size and computation time using Evalica on a synthetic dataset derived from Chatbot Arena as the original dataset was already larger than most existing preference-based NLP datasets. We selected seven dataset sizes, ranging from 10\textsuperscript{1} to 10\textsuperscript{7} pairs, with each size increasing by a factor of ten. For each size, we sampled the required number of pairs with replacement from Chatbot Arena ten times to study the time variance. Computation times were measured using Rust implementations of the methods available in Evalica, and we constructed 95\% confidence intervals using bootstrapping. Figure~\ref{fig:scale} shows that the relationship between dataset size and computation time scales linearly for all methods, indicating good scalability. However, there are clear performance differences for small input sizes, with methods like \citet{Newman:23} being slower initially but converging to similar trends as input size increases. Note that our analysis was limited by the number of models in the version of Chatbot Arena used in our experiments.

\vfill

\section{Conclusion}

We believe that Evalica will foster the creation of reliable and reproducible benchmarks for future NLP systems. We define several potential directions of further work: (1) implementing a larger set of use cases widely used in practice, including confidence interval construction out of the box and additional ranking algorithms, (2) bringing additional performance and memory optimizations, and (3) supporting other popular programming languages with good interoperability with Rust, including JavaScript and Ruby. To the best of our knowledge, Evalica is the first attempt to offer drop-in accelerated preference-based benchmarks, which affects their computational performance and numerical reliability. We expect that a broader adoption of Evalica will result in faster iteration times, more useful experiments, and fewer model-selection mistakes.

\section*{Acknowledgements}

We would like to thank three anonymous reviewers for their useful feedback. We are grateful to the early adopters of Evalica from the Vikhr team \citep{Nikolich:24} and JetBrains~AI team, whose kind feedback allowed improving the library. Note that the correct pronunciation of the library name is \emph{eh-vah-lee-tsah}, which resembles a typical village name in the Balkans.

\clearpage

\bibliography{custom}

\begin{thebibliography}{21}
\providecommand{\natexlab}[1]{#1}

\bibitem[{Abid et~al.(2019)Abid, Abdalla, Abid, Khan, Alfozan, and Zou}]{Abid:19}
Abubakar Abid, Ali Abdalla, Ali Abid, Dawood Khan, Abdulrahman Alfozan, and James~Y. Zou. 2019.
\newblock \href {https://arxiv.org/abs/1906.02569} {{Gradio: Hassle-Free Sharing and Testing of ML Models in the Wild}}.
\newblock \emph{Preprint}, arXiv:1906.02569.

\bibitem[{Bonacich(1987)}]{Bonacich:87}
Phillip Bonacich. 1987.
\newblock \href {https://doi.org/10.1086/228631} {{Power and Centrality: A Family of Measures}}.
\newblock \emph{American Journal of Sociology}, 92(5):1170--1182.

\bibitem[{Bradley and Terry(1952)}]{Bradley:52}
Ralph~Allan Bradley and Milton~E. Terry. 1952.
\newblock \href {https://doi.org/10.2307/2334029} {{Rank Analysis of Incomplete Block Designs: I. The Method of Paired Comparisons}}.
\newblock \emph{Biometrika}, 39(3/4):324--345.

\bibitem[{Brin and Page(1998)}]{Brin:98}
Sergey Brin and Lawrence Page. 1998.
\newblock \href {https://doi.org/10.1016/S0169-7552(98)00110-X} {{The anatomy of a large-scale hypertextual Web search engine}}.
\newblock \emph{Computer Networks and ISDN Systems}, 30(1):107--117.
\newblock Proceedings of the Seventh International World Wide Web Conference.

\bibitem[{Chiang et~al.(2024)Chiang, Zheng, Sheng, Angelopoulos, Li, Li, Zhu, Zhang, Jordan, Gonzalez, and Stoica}]{Chiang:24}
Wei-Lin Chiang, Lianmin Zheng, Ying Sheng, Anastasios~Nikolas Angelopoulos, Tianle Li, Dacheng Li, Banghua Zhu, Hao Zhang, Michael Jordan, Joseph~E. Gonzalez, and Ion Stoica. 2024.
\newblock \href {https://proceedings.mlr.press/v235/chiang24b.html} {{Chatbot Arena: An Open Platform for Evaluating LLMs by Human Preference}}.
\newblock In \emph{Proceedings of the 41st International Conference on Machine Learning}, volume 235 of \emph{Proceedings of Machine Learning Research}, pages 8359--8388. PMLR.

\bibitem[{Dubois et~al.(2024)Dubois, Liang, and Hashimoto}]{Dubois:24}
Yann Dubois, Percy Liang, and Tatsunori Hashimoto. 2024.
\newblock \href {https://openreview.net/forum?id=CybBmzWBX0} {{Length-Controlled AlpacaEval: A Simple Debiasing of Automatic Evaluators}}.
\newblock In \emph{First Conference on Language Modeling}.

\bibitem[{Elo(1978)}]{Elo:78}
Arpad~E. Elo. 1978.
\newblock \emph{{The Rating Of Chess Players, Past \& Present}}.
\newblock Arco Publishing Inc., New York.

\bibitem[{Faggioli et~al.(2024)Faggioli, Dietz, Clarke, Demartini, Hagen, Hauff, Kando, Kanoulas, Potthast, Stein, and Wachsmuth}]{Faggioli:24}
Guglielmo Faggioli, Laura Dietz, Charles L.~A. Clarke, Gianluca Demartini, Matthias Hagen, Claudia Hauff, Noriko Kando, Evangelos Kanoulas, Martin Potthast, Benno Stein, and Henning Wachsmuth. 2024.
\newblock \href {https://doi.org/10.1145/3624730} {{Who Determines What Is Relevant? Humans or AI? Why Not Both?}}
\newblock \emph{Communications of the ACM}, 67(4):31--34.

\bibitem[{Harris et~al.(2020)Harris, Millman, van~der Walt, Gommers et~al.}]{Harris:20}
Charles~R. Harris, K.~Jarrod Millman, St{\'{e}}fan~J. van~der Walt, Ralf Gommers, et~al. 2020.
\newblock \href {https://doi.org/10.1038/s41586-020-2649-2} {{Array programming with NumPy}}.
\newblock \emph{Nature}, 585(7825):357--362.

\bibitem[{Herbrich et~al.(2006)Herbrich, Minka, and Graepel}]{Herbrich:06}
Ralf Herbrich, Tom Minka, and Thore Graepel. 2006.
\newblock \href {https://doi.org/10.7551/mitpress/7503.003.0076} {{TrueSkill\textsuperscript{\texttrademark}: A Bayesian Skill Rating System}}.
\newblock In \emph{Advances in Neural Information Processing Systems~19}, pages 569--576. MIT Press.

\bibitem[{Joshi(2024)}]{Joshi:24}
Vivek Joshi. 2024.
\newblock \href {https://doi.org/10.21105/joss.05901} {{OpenSkill: A faster asymmetric multi-team, multiplayer rating system}}.
\newblock \emph{Journal of Open Source Software}, 9(93):5901.

\bibitem[{Li et~al.(2024)Li, Chiang, Frick, Dunlap, Zhu, Gonzalez, and Stoica}]{Li:24}
Tianle Li, Wei-Lin Chiang, Evan Frick, Lisa Dunlap, Banghua Zhu, Joseph~E. Gonzalez, and Ion Stoica. 2024.
\newblock \href {https://lmsys.org/blog/2024-04-19-arena-hard/} {{From Live Data to High-Quality Benchmarks: The Arena-Hard Pipeline}}.

\bibitem[{MacIver et~al.(2019)MacIver, Hatfield-Dodds et~al.}]{MacIver:19}
David~R. MacIver, Zac Hatfield-Dodds, et~al. 2019.
\newblock \href {https://doi.org/10.21105/joss.01891} {{Hypothesis: A new approach to property-based testing}}.
\newblock \emph{Journal of Open Source Software}, 4(43):1891.

\bibitem[{McKinney(2010)}]{McKinney:10}
Wes McKinney. 2010.
\newblock \href {https://doi.org/10.25080/Majora-92bf1922-00a} {{Data Structures for Statistical Computing in Python}}.
\newblock In \emph{Proceedings of the 9th Python in Science Conference}, SciPy~2010, pages 56--61.

\bibitem[{Newman(2023)}]{Newman:23}
Mark E.~J. Newman. 2023.
\newblock \href {http://jmlr.org/papers/v24/22-1086.html} {{Efficient Computation of Rankings from Pairwise Comparisons}}.
\newblock \emph{Journal of Machine Learning Research}, 24(238):1--25.

\bibitem[{Nikolich et~al.(2024)Nikolich, Korolev, Shelmanov, and Kiselev}]{Nikolich:24}
Aleksandr Nikolich, Konstantin Korolev, Artem Shelmanov, and Igor Kiselev. 2024.
\newblock \href {https://arxiv.org/abs/2405.13929} {{Vikhr: The Family of Open-Source Instruction-Tuned Large Language Models for Russian}}.
\newblock \emph{Preprint}, arXiv:2405.13929.

\bibitem[{Pedregosa et~al.(2011)Pedregosa, Varoquaux, Gramfort, Michel, Thirion, Grisel, Blondel, Prettenhofer, Weiss, Dubourg, Vanderplas, Passos, Cournapeau, Brucher, Perrot, and Duchesnay}]{Pedregosa:11}
Fabian Pedregosa, Ga\"{e}l Varoquaux, Alexandre Gramfort, Vincent Michel, Bertrand Thirion, Olivier Grisel, Mathieu Blondel, Peter Prettenhofer, Ron Weiss, Vincent Dubourg, Jake Vanderplas, Alexandre Passos, David Cournapeau, Matthieu Brucher, Matthieu Perrot, and \'{E}douard Duchesnay. 2011.
\newblock \href {https://jmlr.org/papers/v12/pedregosa11a.html} {{Scikit-learn: Machine Learning in Python}}.
\newblock \emph{Journal of Machine Learning Research}, 12(85):2825--2830.

\bibitem[{Ustalov(2023)}]{Ustalov:23:llmfao}
Dmitry Ustalov. 2023.
\newblock \href {https://doi.org/10.57967/hf/2994} {{Large Language Model Feedback Analysis and Optimization (LLMFAO)}}.
\newblock Dataset.

\bibitem[{Ustalov et~al.(2024)Ustalov, Pavlichenko, and Tseitlin}]{Ustalov:24:crowdkit}
Dmitry Ustalov, Nikita Pavlichenko, and Boris Tseitlin. 2024.
\newblock \href {https://doi.org/10.21105/joss.06227} {{Learning from Crowds with Crowd-Kit}}.
\newblock \emph{Journal of Open Source Software}, 9(96):6227.

\bibitem[{Virtanen et~al.(2020)Virtanen, Gommers, Oliphant, Haberland, Reddy et~al.}]{Virtanen:20}
Pauli Virtanen, Ralf Gommers, Travis~E. Oliphant, Matt Haberland, Tyler Reddy, et~al. 2020.
\newblock \href {https://doi.org/10.1038/s41592-019-0686-2} {{SciPy 1.0: Fundamental Algorithms for Scientific Computing in Python}}.
\newblock \emph{Nature Methods}, 17:261--272.

\bibitem[{Zhou et~al.(2023)Zhou, Lu, Mishra, Brahma, Basu, Luan, Zhou, and Hou}]{Zhou:23}
Jeffrey Zhou, Tianjian Lu, Swaroop Mishra, Siddhartha Brahma, Sujoy Basu, Yi~Luan, Denny Zhou, and Le~Hou. 2023.
\newblock \href {https://arxiv.org/abs/2311.07911} {{Instruction-Following Evaluation for Large Language Models}}.
\newblock \emph{Preprint}, arXiv:2311.07911.

\end{thebibliography}

\clearpage\appendix\onecolumn

\section{\label{sec:usage}Usage Examples}

\begin{listing*}[!ht]
\begin{minted}{pycon}
>>> from evalica import elo, pairwise_frame, Winner
>>> result = elo(
...     xs=["pizza", "burger", "pizza"],
...     ys=["burger", "sushi", "sushi"],
...     winners=[Winner.X, Winner.Y, Winner.Draw],
... )
>>> result.scores
pizza     1014.972058
burger     970.647200
sushi     1014.380742
Name: elo, dtype: float64
>>> df_scores = pairwise_frame(result.scores)
>>> df_scores  # can be used for plotting the pairwise win rate
           pizza     sushi    burger
pizza   0.500000  0.500003  0.501499
sushi   0.499997  0.500000  0.501496
burger  0.498501  0.498504  0.500000
\end{minted}
\caption{\label{lst:elo}An example of computing Elo ranking and the corresponding pairwise win rates with Evalica. Other methods can be applied similarly with a trivial modification: \texttt{bradley\_terry}, \texttt{average\_win\_rate}, etc. See \url{https://github.com/dustalov/evalica/blob/master/Tutorial.ipynb} for an executable example.}
\end{listing*}

\begin{listing*}[!ht]
\begin{minted}{python}
# index the compared models to save time by not re-indexing them at each round
*_, index = evalica.indexing(
    xs=df["model_a"],  # series with model A identifiers
    ys=df["model_b"],  # series with model B identifiers
)

bootstrap: list["pd.Series[str]"] = []  # assuming model names are strings

for r in range(BOOTSTRAP_ROUNDS):
    # for reproducibility, set the random seed equal to the number
    # of the bootstrapping round
    df_sample = df_arena.sample(frac=1.0, replace=True, random_state=r)

    # estimate the Bradley-Terry scores for the given sample
    result_sample = evalica.bradley_terry(
        xs=df_sample["model_a"],
        ys=df_sample["model_b"],
        winners=df_sample["winner"],
        index=index  # use the index built above to speed up
    )

    bootstrap.append(result_sample.scores)

# this is a data frame with BOOTSTRAP_ROUNDS rows,
# each row represents the score of each model at the r-th round
df_bootstrap = pd.DataFrame(bootstrap)

# this is a data frame with confidence intervals of scores
# for each compared model
df_bootstrap_ci = pd.DataFrame({
    "lower": df_bootstrap.quantile(.025),
    "rating": df_bootstrap.quantile(.5),
    "upper": df_bootstrap.quantile(.975),
}).reset_index(names="model").sort_values("rating", ascending=False)
\end{minted}
\caption{\label{lst:bootstrap}An example of bootstrapping a 95\% confidence interval of \citet{Bradley:52} scores with Evalica and pandas \citep{McKinney:10}. Any other supported model can be applied after a trivial modification. For simplicity, we do not show an example with \texttt{scipy.stats.bootstrap} \citep{Virtanen:20}, yet it is possible. See \url{https://github.com/dustalov/evalica/blob/master/Chatbot-Arena.ipynb} for an executable example.}
\end{listing*}

\clearpage

\begin{figure*}[!ht]
  \includegraphics[width=\textwidth]{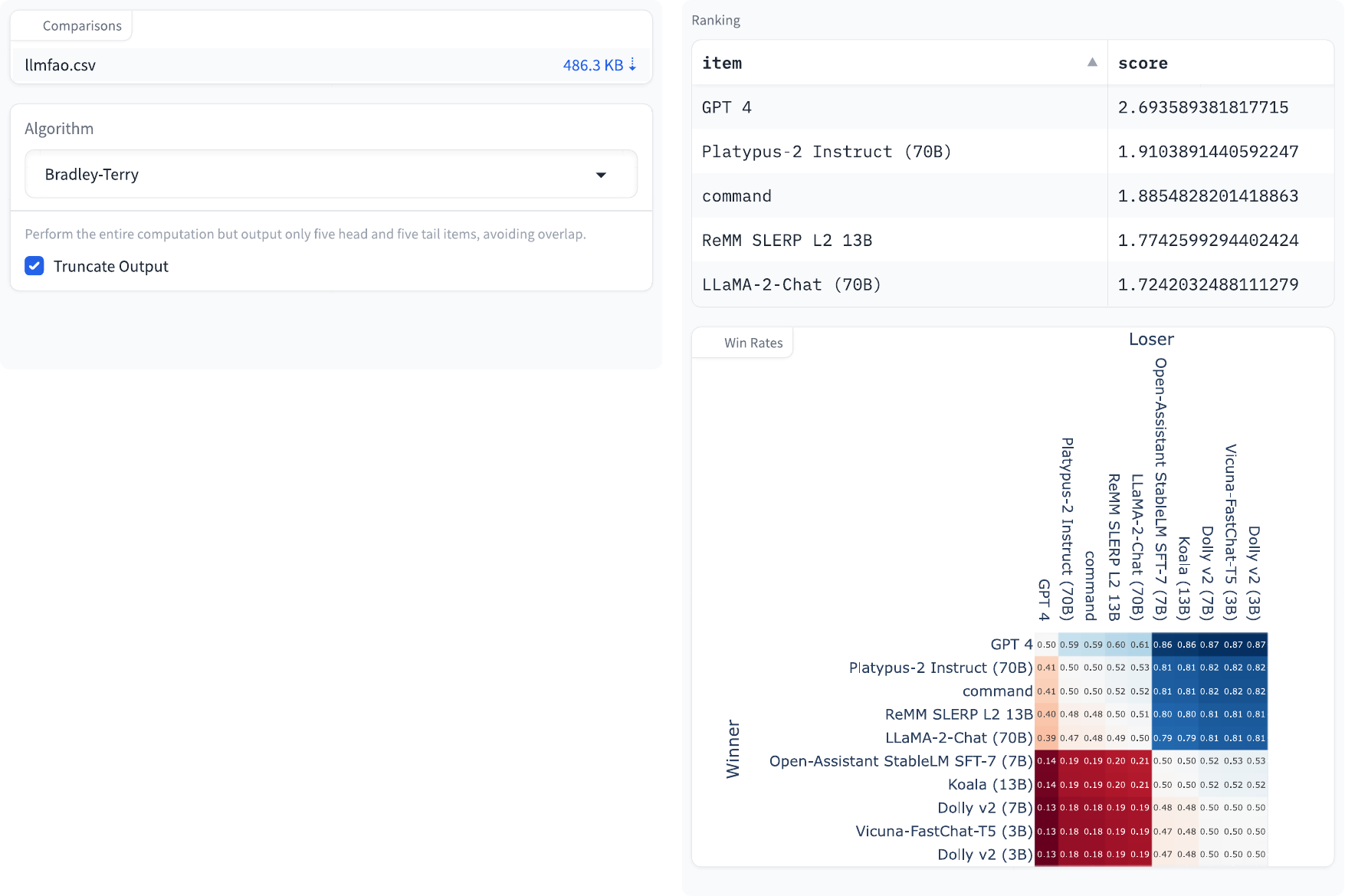}
  \caption{\label{fig:gradio}A screenshot of the Evalica's Web interface with the LLMFAO benchmark \citep{Ustalov:23:llmfao}. On the left, there are the input file, algorithm choice, and additional parameters. On the right, there is a table with the ranking results and a win rate plot. For the sake of brevity, we showed only a truncated output, with no columns corresponding to the number of compared pairs and the current rank of the model. A live example can be accessed at \url{https://huggingface.co/spaces/dustalov/pair2rank}.}
\end{figure*}

\begin{figure*}[!ht]
\begin{minted}{console}
$ head -n6 food.csv | column -ts,
left    right  winner
Pizza   Sushi  left
Burger  Pasta  right
Tacos   Pizza  left
Sushi   Tacos  right
Burger  Pizza  left
$ evalica -i food.csv bradley-terry | column -ts,
item    score               rank
Tacos   2.509025136024378   1
Sushi   1.1011561298265815  2
Burger  0.8549063627182466  3
Pasta   0.7403814336665869  4
Pizza   0.5718366915548537  5
\end{minted}
\caption{\label{fig:cli}An example of using a command-line interface of Evalica to process a file in the comma-separated values format and print the item ranks and estimated scores.}
\end{figure*}

\end{document}